\documentclass[10pt,twocolumn,letterpaper]{article}

\usepackage{iccv}              

\definecolor{iccvblue}{rgb}{0.21,0.49,0.74}
\usepackage[pagebackref,breaklinks,colorlinks,allcolors=iccvblue]{hyperref}

\usepackage[export]{adjustbox}
\usepackage{multirow,multicol}
\usepackage{colortbl}
\usepackage{makecell}
\definecolor{mygray}{gray}{0.6}

\def\paperID{10544} 
\def\confName{ICCV}
\def\confYear{2025}

\title{CLIP-GS: Unifying Vision-Language Representation with 3D Gaussian Splatting}

\author{%
  Siyu Jiao$^{1,2}$ \and Haoye Dong$^{3}$ \and Yuyang Yin$^{1,2}$ \and Zequn Jie$^{4}$ \and Yinlong Qian$^{4}$ \and Yao Zhao$^{1,2\dag}$  ~~~~~~ Humphrey Shi$^{5,6}$ ~~~~~~ Yunchao Wei$^{1,2\dag}$ \\
  \\
  $^{1}$ Institute of Information Science, Beijing Jiaotong University\\
  $^{2}$ Visual Intellgence +X International Cooperation Joint Laboratory of MOE \\
  $^{3}$ National University of Singapore ~~~~~~ $^{4}$ Meituan \\
   $^{5}$ Georgia Institute of Technology ~~~~~~ $^{6}$ Picsart AI Research (PAIR)
}

\begin{document}


\twocolumn[{
    \renewcommand\twocolumn[1][]{#1}
    \maketitle
    \vspace{-1.1cm}
    \begin{center}
        \centering\captionsetup{type=figure}
        \includegraphics[width=0.9\linewidth]{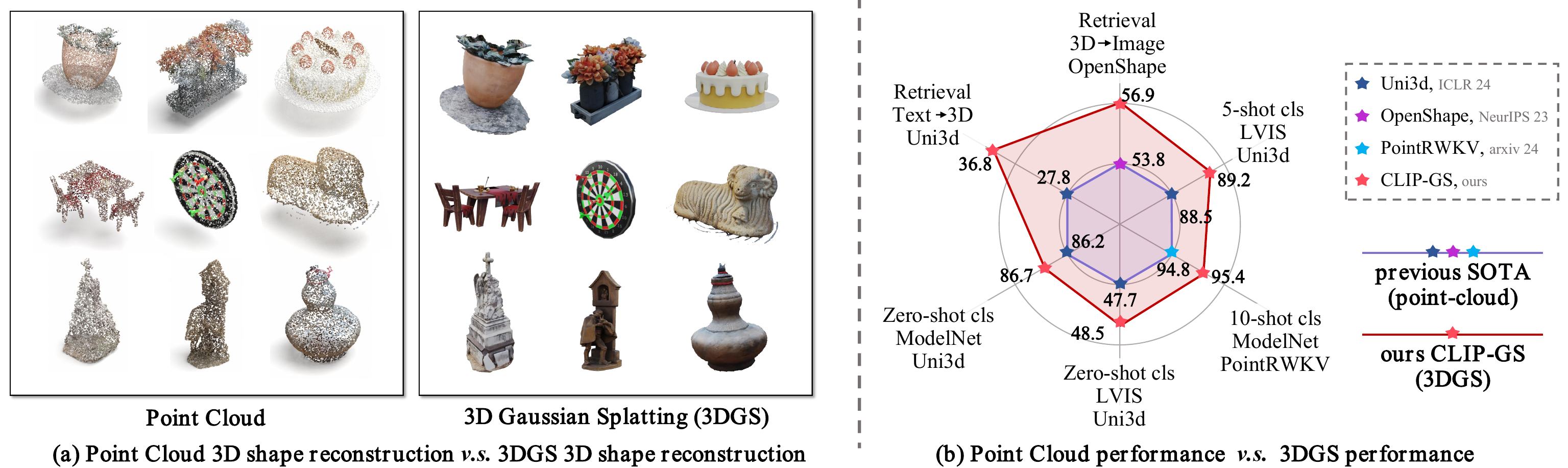}
        \vspace{-2mm}
        \captionof{figure}{(a) Comparison between point cloud reconstruction and 3D Gaussian Splatting (3DGS) reconstruction. (b) The 3DGS approach outperforms point cloud methods across multiple 3D perception tasks, indicating its superior 3D object representation capabilities. These results suggest that 3D perception based on 3DGS holds significant advantages over point cloud-based methods.} 
        \label{fig:abs}
    \end{center}
}]

\def\thefootnote{\dag}\footnotetext{Corresponding author}

\begin{abstract}
Recent works in 3D multimodal learning have made remarkable progress.
However, typically 3D multimodal models are only capable of handling point clouds. 
Compared to the emerging 3D representation technique, 3D Gaussian Splatting (3DGS), the spatially sparse point cloud cannot depict the texture information of 3D objects, resulting in inferior reconstruction capabilities. This limitation constrains the potential of point cloud-based 3D multimodal representation learning.
In this paper, we present CLIP-GS, a novel multimodal representation learning framework grounded in 3DGS. 
We introduce the GS Tokenizer to generate serialized gaussian tokens, which are then processed through transformer layers pre-initialized with weights from point cloud models, resulting in the 3DGS embeddings.
CLIP-GS leverages contrastive loss between 3DGS and the visual-text embeddings of CLIP, and we introduce an image voting loss to guide the directionality and convergence of gradient optimization.
Furthermore, we develop an efficient way to generate triplets of 3DGS, images, and text,  facilitating CLIP-GS in learning unified multimodal representations.
Leveraging the well-aligned multimodal representations, CLIP-GS demonstrates versatility and outperforms point cloud-based models on various 3D tasks, including multimodal retrieval, zero-shot, and few-shot classification.


\end{abstract}
\vspace{-2mm}
    
\section{Introduction}
\label{sec:intro}
 
Learning 3D representations stands as the most popular basic topic in 3D computer vision, driven by the increasing demand for real-world applications in autonomous driving \cite{yan2024street, zhou2024drivinggaussian}, and embodied AI \cite{embodiedscan, huang2023embodied, huang2023voxposer, xu2024esam, habitat19iccv, zhu2024unifying, geng2022gapartnet}. 
Existing works in 3D representation learning have made remarkable progress, particularly through the development of transformer-based approaches \cite{guo2021pct, yan2020pointasnl, yu2022pointbert, pan20213d, lu20223dctn}, as well as mamba-based approaches \cite{zhang2024point, liang2024pointmamba, han2024mamba3d, li20243dmambacomplete}. 

Pre-trained vision-language models have rapidly developed in the 2D domain in the past few years. Notable examples \cite{clip, slip, siglip, evaclip} achieving image-text alignment through contrastive learning techniques supported by large-scale 2D datasets. Following the advent of large-scale 3D datasets Objaverse \cite{objaverse, objaversexl}, a few works \cite{ulip, ulip2, uni3d, openshape} explore multimodal pre-training within the 3D domain. 
However, the emphasis predominantly remains on point cloud, which, as a sparse spatial representation, offers limited 3D reconstruction capabilities compared to the emerging 3D modeling method 3D Gaussian Splatting (3DGS), as shown in Fig. \ref{fig:abs} (a). Therefore, enhancing 3D perception via 3DGS models has become an urgent challenge to address.

In this work, we introduce a multimodal representation learning method leveraging 3DGS, termed \textbf{CLIP-GS}. We outline the design of a 3D encoder processing 3DGS inputs while aligning with CLIP's visual and textual representations. Within CLIP-GS, we initially employ FPS \& kNN to form 3DGS into gaussian patches. Subsequently, we design a GS Tokenizer to yield serialized gaussian tokens. These tokens are then processed by a series of transformer layers, pre-trained on point-cloud data, to generate distinct gaussian features. 
To address the challenge of varying viewpoints in different rendered images, we introduce an image voting loss mechanism. This mechanism, grounded in the CLIP-facilitated congruency between images and text, employs a \textit{voting} strategy to guide the directionality and convergence of gradient optimization.


Apart from the architectural design, the limited availability of 3DGS poses a significant challenge. To counter this, we explore an efficient approach for generating 3DGS, resulting in $\sim$ 240K 3DGS from the Objaverse dataset \cite{objaverse, objaversexl}. By leveraging the weights from a pre-trained multimodal point cloud model, CLIP-GS is capable of learning well-aligned multimodal representations with only 240K samples. 
CLIP-GS demonstrates strong adaptability to various downstream 3D perception tasks, consistently outperforming point cloud-based models, as illustrated in Fig. \ref{fig:abs} (b).

We evaluate our CLIP-GS across three fundamental 3D tasks: multimodal retrieval, zero-shot classification, and few-shot classification. Extensive experiments show that CLIP-GS is highly effective across diverse 3D tasks. CLIP-GS achieves superior performance on Text $\rightarrow$ 3D retrieval (27.8\% $\rightarrow$ 36.8\%) and 3D $\rightarrow$ Image retrieval (53.8\% $\rightarrow$ 56.9\%) in terms of R@1. CLIP-GS enhances the performance of Objaverse-GS and ModelNet-GS datasets by +0.8\%, +0.5\% in zero-shot classification, and +0.6\%, +0.4\% in few-shot classification. 
Remarkably, our approach outperforms the existing point cloud-based models, and establishes new state-of-the-art results on all benchmarks.

Overall, our contributions are summarized as follows:
\begin{itemize}[itemsep=2pt,topsep=0pt,parsep=0pt]
\item We propose CLIP-GS, a simple yet effective framework for encoding 3DGS into features, leveraging a contrastive learning paradigm for multimodal per-taining.
\item We develop an efficient method for generating triplets comprising 3DGS, rendered images, and text to enable CLIP-GS to learn unified multimodal representations.
\item Extensive experiments across various downstream tasks demonstrate that our method consistently outperforms others in all tasks, including zero-shot 3D classification, few-shot 3D classification, and multimodal retrieval. 
\end{itemize}


\section{Related Work}
\label{sec:related}

\noindent \textbf{Multimodal Representation Learning.}
Multimodal representation learning aims to align the feature representations of various modalities into a unified feature space. 
Most research works focus on learning representations of image and language modalities. 
Transformer-based frameworks \cite{jia2021scaling, clip, evaclip, siglip, slip, sun2024alpha, maft, maftp, zhang2023controlvideo} predominate this space, facilitating the learning of interactions between image and language. The paradigm typically employs a contrastive learning paradigm and relies on extensive pre-training using substantial sets of image-caption pairs.

Recent works explore applying image-text pretraining models \cite{clip} to the 3D domain. These approaches typically conform to one of two primary paradigms. The first \cite{zhu2023pointclip, zhang2022pointclip, clip2} projects point clouds into 2D images, leveraging 2D vision-language models for inference. However, multi-view rendering and 2D image inference incur substantial computational overhead, and 3D spatial details risk omission during projection. The second paradigm trains a native 3D point cloud encoder \cite{ulip, ulip2, openshape, uni3d}, with the aim of aligning 3D shape representations with the feature space of vision-language models, \textit{e.g.} CLIP.

However, a limitation in current 3D multimodal representation learning is the prevailing use of point clouds as the 3D input. The inherently sparse and discrete nature of point clouds inhibits their capacity to articulate textural information, thereby impeding the fidelity of 3D reconstruction capabilities. This property constrains the potential upper bound of 3D multimodal representation learning.

\noindent \textbf{3D Representations.}
3D representation methods have significantly advanced over time. Polygonal meshes \cite{lorensen1998marching,hoppe1992surface} represent surfaces using edges, vertices, and faces. Point cloud, on the other hand, provides an unstructured 3D data format composed of position attributes, optionally including color or intensity. Point cloud is widely used in 3D perception tasks \cite{qi2017pointnet,qi2017pointnet++,kazhdan2006poisson} like classification, detection, and segmentation. However, the intrinsic limitations in reconstruction quality constrain the potential of 3D perception tasks.
NeRF \cite{mildenhall2021nerf} addresses the reconstruction quality problem by using MLP layers to represent 3D content implicitly. Subsequent works \cite{muller2022instant, barron2021mip} significantly improved training efficiency and rendering quality. However, Nerf struggles with rendering speed since millions of queries to the MLP network, and its implicit representation poses difficulties for widespread implementation in 3D perception models.

Recently, 3D Gaussian Splatting (3DGS) \cite{kerbl20233d} uses explicit 3D gaussian points to represent objects and scenes, with each gaussian point characterized by position, rotation, scale, color, and opacity attributes. 3DGS offers improved spatial accuracy and efficiency in capturing geometric shapes. Due to these advantages, many downstream tasks have increasingly adopted 3DGS as a primary 3D representation, achieving remarkable performance in segmentation \cite{hu2024semantic, ye2025gaussian}, generation \cite{tang2023dreamgaussian, zhong2024dreamlcm, yin20234dgen, liang2024diffusion4d}, autonomous driving \cite{yan2024street, zhou2024drivinggaussian}, .etc. Consequently, investigating 3D representation learning using 3DGS is both valuable and instrumental in advancing progress across these applications.

\section{Scalable Triplet Generation for 3DGS}
\label{sec:data}

CLIP-GS learns a unified representation space of images, text, and 3DGS via pre-training on triplets from these three modalities. This section outlines the method deployed for generating the triplets required for pre-training.

\begin{figure}[h]
    \centering
    \begin{subfigure}{.25\textwidth}
        \centering
        \renewcommand\arraystretch{1.3} 
        \begin{tabular}{c|c}
            \hline
            \textbf{Modality} &  \textbf{Scale} \\
            \hline
            3D models & $\sim$ 240K \\
            Images & $\sim$ 8.6M \\
            Text Descriptions & $\sim$ 240K \\
            \hline
        \end{tabular}
        \caption{Data scale}
        \label{fig:statistics1}
    \end{subfigure}%
    \hspace{2mm}
    \begin{subfigure}{.21\textwidth}
        \centering
        \includegraphics[width=.9\linewidth]{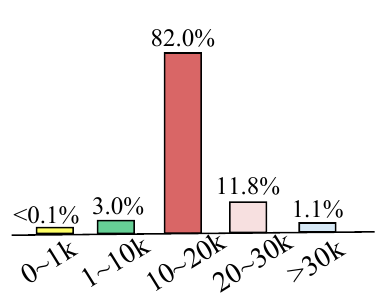}
        \caption{Number of Gaussian points.}
        \label{fig:statistics2}
    \end{subfigure}
    \caption{Statistics of 3DGS Triplets.}
    \label{fig:statistics}
\end{figure}

\noindent \textbf{3D shapes collection.}
Our triplet is constructed using Objaverse \cite{objaverse} and Objaverse-XL \cite{objaversexl}, the largest-scale realistic 3D dataset. Following the filtering criteria in \cite{cap3d}, we exclude overly simple or meaningless 3D shapes, selecting $\sim$ 240K high-quality 3D shapes. These models showcase a rich diversity of colors and textures, effectively capitalizing on the representational strength of 3DGS. The scale statistics of three modalities are shown in Fig. \ref{fig:statistics1}. In the supplementary materials, we provide visualizations of both the excluded and retained data.

\noindent \textbf{Rendering images and captions for 3D shapes.}
For each 3D shape, we employ Blender \cite{blender} to render 36 distinct images by randomly selecting viewpoints within the spatial domain, scaling each image to a resolution of 512 $\times$ 512, and centering the 3D shape within the image.
Each 3D shape is paired with a descriptive caption provided by Cap3D \cite{cap3d}. Contrary to the approaches \cite{ulip2, uni3d, openshape} assign separate captions to each image from different viewpoints, Cap3D offers a unifying caption that encapsulates the essence of the 3D shape. This approach minimizes misalignment issues that can arise from inconsistent captions associated with images taken from various viewpoints.

\noindent \textbf{Generation of 3DGS.}
The generation of 3DGS models is facilitated through the vanilla 3DGS reconstruction algorithm \cite{kerbl20233d}. To mitigate the demands of optimization and storage costs, we streamline the spherical harmonics (SH) functions that delineate the color attributes of the 3DGS, setting the SH degree to 0. \textit{i.e.} one gaussian point retains a singular color variant (R,G,B). This simplification results in a significant reduction in the storage demands of the 3DGS models. Additionally, the position and color attributes of 3DGS are initialized via point clouds, enabling the completion of optimization within 5,000 iterations. We depict the distribution of the number of gaussian points in Fig. \ref{fig:statistics2}, where most 3DGS contain 10$\sim$20k gaussian points. Sec. \ref{sec:abs} encompasses an ablation study on the impact of iterations and SH degrees, proving our approach's ability to effectively reduce the burden of model construction and storage while maintaining reconstruction quality.

\begin{figure*}[t]
\begin{center}
\vspace{-5mm}
   \includegraphics[width=0.99\linewidth]{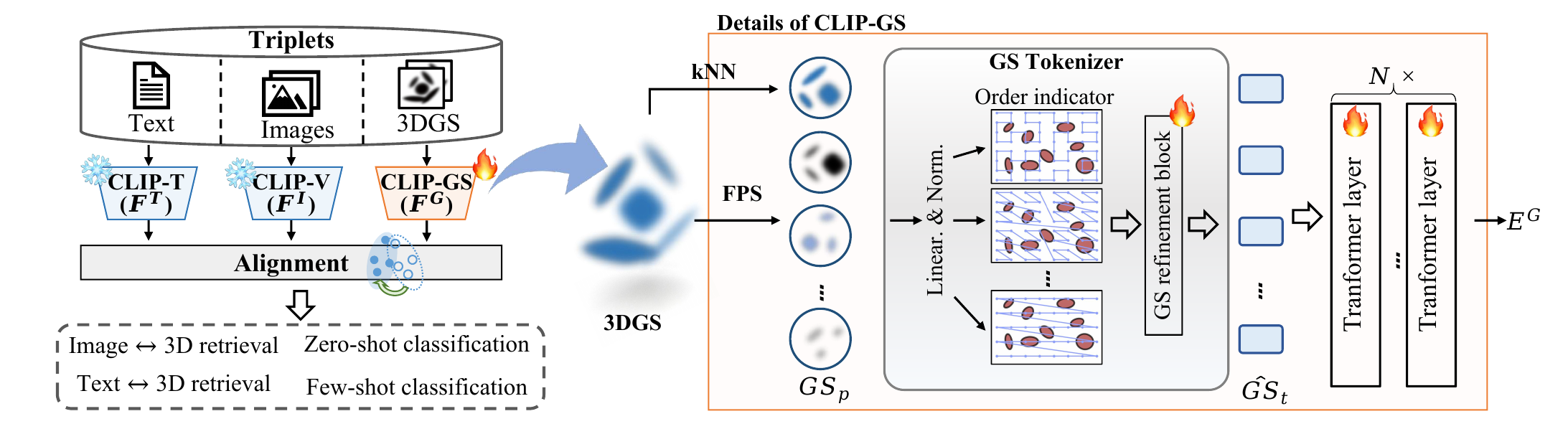}
\end{center}
\vspace{-5mm}
   \caption{
    Overview of the CLIP-GS. Within CLIP-GS, the FPS \& kNN is first used to form gaussian patches. Then, we design the GS Tokenizer to obtain the serialized gaussian tokens. Finally, the entire sequence of Gaussian tokens is processed by a series of transformer layers that have been pre-trained on point clouds, resulting in the Gaussian features.
   }
\label{fig:framework}
\end{figure*}

\section{Methodology}
\label{sec:method}

We present CLIP-GS,  a unified 3D pretraining framework for large-scale 3D representation learning by aligning 3DGS embeddings with the text-image aligned embeddings. The complete framework is shown in Fig.~\ref{fig:framework}.
We introduce the feature extraction process from 3DGS, detailed in Sec. \ref{sec:feature-extraction}. Following this, we outline the training pipeline for CLIP-GS in Sec. \ref{sec:multi-align}.
The model training leverages contrastive learning, which is applied to the gaussian embeddings with the visual and textual embeddings of CLIP. We also introduce a novel loss function, termed \textit{image voting loss}, to guide the convergence of gradient optimization.

\subsection{Feature Extraction}
\label{sec:feature-extraction}

Within CLIP-GS, the FPS \& kNN is first used to form 3DGS into gaussian patches. Then, we design the GS Tokenizer to generate the serialized gaussian tokens. Finally, the sequence of gaussian tokens is processed by a series of transformer layers pre-trained on point cloud data, resulting in the gaussian embeddings. 

\noindent \textbf{Forming GS patches.}  
In 3DGS, each gaussian point is defined by a set of attributes: position $P \in \mathbb{R}^3$, the color represented via a spherical harmonic function $C \in \mathbb{R}^{3(sh+1)^2}$, opacity $\alpha \in \mathbb{R}^1$, scale $S \in \mathbb{R}^3$, and rotation $R \in \mathbb{R}^4$. We set $sh=0$ in $C$, effectively compressing the gaussian point into a vector in $\mathbb{R}^{14}$ . 
Given a 3DGS $GS$, Farthest Point Sampling (FPS) and k-Nearest Neighbor (kNN) algorithms are adopted to form $GS$ into local patches $GS_p \in \mathbb{R}^{g \times n \times 14}$, $g$ and $n$ denote the number of patches and the number of neighbors for each patch, respectively.

\begin{figure}[ht]
\begin{center}
   \includegraphics[width=0.99\linewidth]{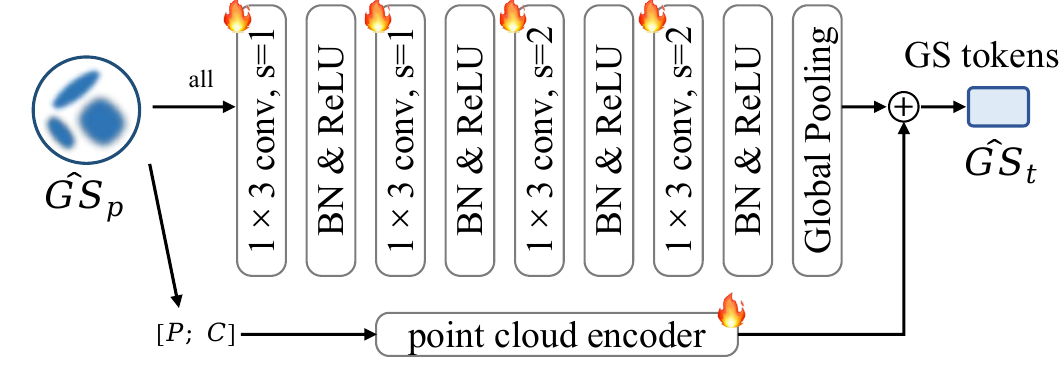}
\end{center}
\vspace{-6mm}
   \caption{
    Details of GS refinement block.
   }
\label{fig:conv1d}
\end{figure}

\noindent \textbf{GS Tokenizer.}
The GS Tokenizer aims to derive the serialized gaussian tokens from gaussian patches. Within GS Tokenizer, $\mathrm{Sigmoid}$ function is first employed to normalize the opacity ($\alpha$) and scaling attributes ($S$)  within a uniform range, and linearize the rotation attribute $R$ into a $\mathbb{R}^{3 \times 3}$ rotation matrix. 
Next, a multi-way ordering strategy, including \textit{xyz}-order, Hilbert curve \cite{hilbert1935stetige}, and Z-order, is adopted to reorganize the $GS_p$, resulting in ordered gaussian patches $\hat{GS_p}$. 
We then design the GS refinement block to extract GS tokens, details in Fig. \ref{fig:conv1d}. 
Here, position and color attributes ($P$ \& $C$) are extracted and input into a point cloud encoder, as detailed in \cite{uni3d}.
Simultaneously, 1 $\times$ 3 convolution layers with batch normalization (BN) and ReLU activation are applied to distill gaussian features. The outputs of these processes are finally fused to obtain GS tokens $\hat{GS_t} \in \mathbb{R}^{g \times d}$, where $d$ denotes the dimension of GS tokens.

\noindent \textbf{Transformer layers.}
We utilize the standard Transformer to obtain 3DGS embeddings $E^G$, a structure equivalent to the 2D Vision Transformer (ViT) \cite{vit}. We reuse the multimodal pre-trained point cloud weights from \cite{uni3d}. Experimental results indicate that these point cloud pre-trained weights offer substantial advantages compared to transformer weights pre-trained on 2D images.

\subsection{Multi-model Alignment}
\label{sec:multi-align}

CLIP-GS aligns the triplet composed of 3DGS, 2D rendered images, and text descriptions into a unified feature space. This alignment process follows the point cloud pretaining approaches \cite{uni3d, ulip, ulip2, openshape}, as depicted in Fig. \ref{fig:framework} (left). The pre-trained vision language model EVA-CLIP \cite{evaclip} is adopted during training. 

\noindent \textbf{Triplet alignment target.}
We denote our CLIP-GS as $F^G$, and the text and image encoders in EVA-CLIP as $F^T$ and $F^I$, respectively. $F^G$ is trained to learn 3D representations by aligning them with established 2D vision and text representations. Both $F^T$ and $F^I$ are frozen since they are well-aligned, leaving only $F^G$ as the learnable component. 
Consider a batch of $N$ triplets $\{(G_i, I_i, T_i)\}_{i=1}^N$, where $G_i$, $I_i$ , $T_i$ respectively represent a 3DGS model and its corresponding rendering image and text description. 
We first extract the image embeddings $E^I_i = F^I(I_i)$, and text embeddings $E^T_i = F^T(T_i)$ using the pre-trained $F^I$ and $F^T$ in EVA-CLIP. The normalized embeddings for the sampled triplets ($\{\hat{E^G_i}, \hat{E^T_i}, \hat{E^I_i}\}_{i=1}^N$) are computed as: 
$\{\hat{E^G_i}=\frac{E^G_i}{|E^G_i|}, \hat{E^T_i}=\frac{E^T_i}{|E^T_i|}, \hat{E^I_i}=\frac{E^I_i}{|E^I_i|}\}_{i=1}^N$. 
The triplet alignment target is to align $\hat{E^G_i}$ with $\hat{E^T_i}$ and $\hat{E^I_i}$.

\noindent \textbf{3D-Text objective.}
Within the triplets, a one-to-one correspondence exists between the 3DGS and the text description. To align the 3DGS with the text description, we employ a contrastive loss function, $\mathcal{L}_{\text{text}}$: 
\begin{equation}
\footnotesize
\mathcal{L}_{\text{text}} = -\frac{1}{2N} \sum_{i=1}^N (\mathrm{Contra}(E^G_i, E^T) + \mathrm{Contra}(E^T_i, E^G))
\label{eq:ltext}
\end{equation}
\begin{equation}
\footnotesize
\mathrm{Contra}(A_i, B) = \log\frac{\exp(A_i\cdot B_i/\tau)}{\sum_{j} \exp(A_i\cdot B_j/\tau)}, B = \{B_i\}_{i=1}^N
\label{eq:lcontra}
\end{equation}

where $\tau$ is a learnable temperature parameter employed to control the scale of the embeddings.

\noindent \textbf{3D-Image objective.}
The relationship between 3DGS and rendered images is characterized by a one-to-many correspondence. Existing point cloud pre-training approaches randomly sample one image from the rendering set and impose a contrastive loss restriction, similar to Eq. (\ref{eq:ltext}) \& (\ref{eq:lcontra}). However, varying camera angles may yield significantly different image features, leading to suboptimal optimization results. In response, we propose the \textit{image voting loss} ($\mathcal{L}_{\text{img}}$).
We sample $K$ images and obtain $K$ normalized image embeddings ($\{E^I_{i,k}\}_{k=1}^K$) for one 3DGS.
$\mathcal{L}_{img}$ utilizes the pretrained EVA-CLIP to calculate a voting score ($S$) between $\{E^I_{i,k}\}_{k=1}^K$ and $\hat{E^T_i}$, which scores the set of rendered images. $S$ is formulated as:
\begin{equation}
\footnotesize
S_i = \frac{\hat{E^T_i} \cdot E^I_{i,k}}{|\hat{E^T_i}| \cdot |E^I_{i,k}|}, k = \{1,2...K\}
\label{eq:lmcontra}
\end{equation}

In the default setting, we set $k=5$ for each 3DGS, \textit{i.e.} we randomly select rendered images from five different viewpoints to form a batch.
The contribution of each image to $\mathcal{L}_{\text{img}}$ is then determined using a voting mechanism, as described by the following formula:
\begin{equation}
\footnotesize
\mathcal{L}_{\text{img}} = -\frac{1}{2N} \sum_{i=1}^N  S_i \cdot (\mathrm{Contra}(E^G_i, E^I) + \mathrm{Contra}(E^I_i, E^G))
\end{equation}

Finally, we minimize $\mathcal{L}_{\text{text}}$ and $\mathcal{L}_{\text{img}}$ for all modality pairs:

\begin{equation}
\mathcal{L} = \mathcal{L}_{\text{text}} + \mathcal{L}_{\text{img}}
\end{equation}

\section{Experiments}
\label{sec:exp}
We conduct experiments on numerous typically 3D tasks to validate the multimodal performance of the proposed CLIP-GS, including multimodal retrieval (Sec. \ref{sec:ovr}), zero-shot classification (Sec. \ref{sec:zs}) and few-shot classification (Sec. \ref{sec:fs}). Furthermore, we perform ablation studies (Sec. \ref{sec:abs}) on CLIP-GS to verify the model efficacy, and explore the effects of scaling up the model size of CLIP-GS (Sec. \ref{sec:scaling}).

\subsection{Multimodal Retrieval}
\label{sec:ovr}

  

     

  

\begin{table*}[t]
\scriptsize
  \setlength{\tabcolsep}{5pt}
  \centering
  
  \resizebox{0.9\textwidth}{!}{
    \begin{tabular}{lcccccccc}
    \toprule
    \multirow{2}[2]{*}{} & \multirow{2}[2]{*}{Modality} & \multirow{2}[2]{*}{3D repr}  & R@1  & R@5  & R@10  & R@1  & R@5  & R@10  \\
\cmidrule{4-9}          &&  & \multicolumn{3}{c}{Text $\rightarrow$ 3D} & \multicolumn{3}{c}{3D $\rightarrow$ Text}\\
    \midrule
    ULIP 2 \cite{ulip2} & \multirow{3}[5]{*}{Text \& 3D} & PC & 4.5 & 14.7 & 23.1  & 5.3      & 16.8      & 25.9 \\
    OpenShape-SparseConv \cite{openshape}  &  & PC & 22.6&	50.2&	64.3& 20.1 & 46.2&	60.4\\
     OpenShape-PointBERT \cite{openshape} & & PC & 24.4&	52.7&	66.0&	22.6&	49.6&	63.5 \\
     Uni3D \cite{uni3d} & & PC & 27.8&	57.0&	70.3&	23.1&	49.5	&62.4\\
     \textbf{CLIP-GS (ours)} & & 3DGS & \textbf{36.8}	& \textbf{68.1}	& \textbf{79.9}	& \textbf{30.0}	& \textbf{59.1}	& \textbf{71.3}\\
    \midrule
    & & & \multicolumn{3}{c}{Image $\rightarrow$ 3D} & \multicolumn{3}{c}{3D $\rightarrow$ Image}\\ \cmidrule{4-9}
    ULIP 2 \cite{ulip2} & \multirow{3}[3]{*}{Image \& 3D} & PC & 5.6      & 15.3 & 21.8 & 25.0 & 50.0& 62.1 \\ 
    OpenShape-SparseConv \cite{openshape} &       & PC &59.8&	85.3&	92.1&	49.5&	77.2&	86.1 \\
    OpenShape-PointBERT \cite{openshape} & & PC & 61.6&	86.4&	92.7&	53.8&	80.7&	88.6\\
     Uni3D \cite{uni3d} & & PC & 65.1	&88.4	&93.9	&49.3&	75.7&	84.5 \\
     
     \textbf{CLIP-GS (ours)} & & 3DGS & \textbf{75.6}	& \textbf{93.9}	& \textbf{97.1}	& \textbf{56.9}	& \textbf{82.3}	& \textbf{89.3} \\

    \bottomrule
    \end{tabular}
    }
  \caption{Multimodal retrieval on Objaverse-GS. For a fair comparison, all methods are trained without Objaverse-LVIS shapes. 3D repr denotes the form of 3D shapes representation. Our CLIP-GS employs 3DGS, while prior works \cite{ulip2, openshape, uni3d} use point clouds (PC).}
  \label{tab:ovr}%
  
\end{table*}%

\begin{figure*}[t]
\begin{center}
   \includegraphics[width=0.99\linewidth]{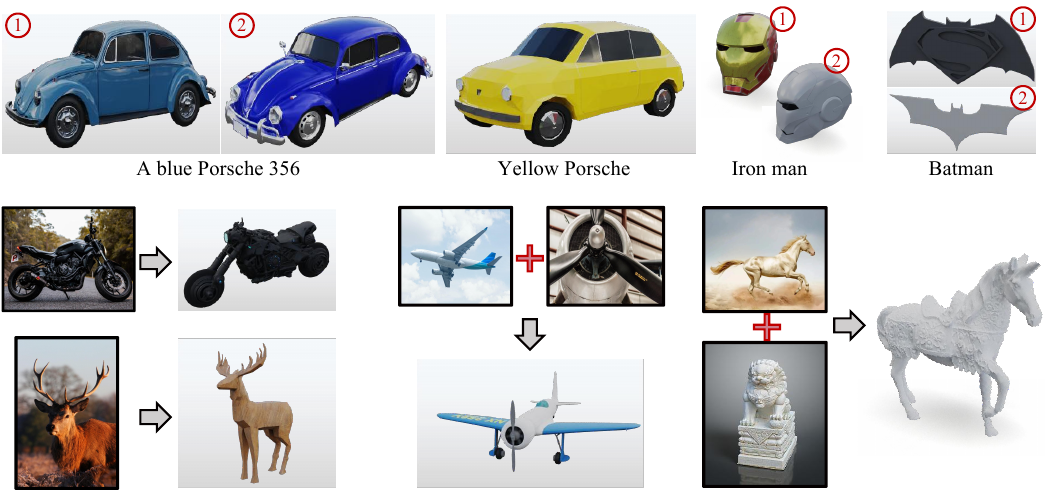}
\end{center}
   \caption{
    Image / text $\rightarrow$ 3D shape retrieval results. Top: we query the most similar or top 2 similar 3D shapes for each text. Bottom: we take one or two images as inputs and retrieve the most similar 3D shape.
   }
\label{fig:visr}
\end{figure*}

We begin our evaluation with the multimodal retrieval task.

\noindent \textbf{Dataset.}
We sample 20,000 3D shapes from Objaverse-LVIS \cite{objaverse}, an annotated subset of Objaverse that includes 1,156 LVIS categories. Utilizing the method outlined in Sec. \ref{sec:data}, we create triplets of rendering images, text descriptions, and 3DGS for each 3D shape, forming a dataset referred to as Objaverse-GS. Objaverse-GS can be applied to tasks such as multimodal retrieval, as well as zero-shot 3D classification and few-shot 3D classification.

\noindent \textbf{Evaluation metrics.}
We evaluate retrieval tasks by using 3D shapes to retrieve corresponding images/text or using images/text to retrieve 3D shapes.
Following by \cite{wang2024semantics,wang2024cl2cm,ren2021learning,li2019visual}, we measure the performance by $R@K$ (Recall at K) defined as the fraction of queries for which the correct item is retrieved in the closest $K$ points to the query.
$K$ is set to \{1, 5, 10\} respectively. 

\noindent \textbf{Comparisons with state-of-the-art methods.}
We compare CLIP-GS with the previous excellent multimodal point-cloud approaches \cite{ulip2, uni3d, openshape}. In CLIP-GS, 3DGS is adopted for 3D shape retrieval, whereas other methods use point clouds for their retrieval processes. 
The results are shown in Tab. \ref{tab:ovr}. Our CLIP-GS outperforms point cloud-based methods across all retrieval tasks by a large margin.
Compared to the previously best-performing approaches, CLIP-GS achieves improvements of +9.0, +6.9, +10.5, and +7.6 in terms of $R@K$ for Text $\rightarrow$ 3D, 3D $\rightarrow$ Text, Image $\rightarrow$ 3D, and 3D $\rightarrow$ Image retrieval tasks, respectively.

\noindent \textbf{Qualitative analysis.}
In Fig. \ref{fig:visr}, we showcase how CLIP-GS successfully retrieves 3D shapes from text or real-world images. By computing the cosine-similarity between images/text and 3D shapes, we retrieve the most similar or the Top2 similar 3D shapes. The results indicate that CLIP-GS retrieves reasonable 3D shapes based on text in the query set (Fig. \ref{fig:visr} top). 
CLIP-GS performs well when retrieving real-world images. In addition, we explore taking two images as input to retrieve 3D shapes similar to both images (Fig. \ref{fig:visr} bottom). It is evident that CLIP-GS has learned the encoding of 3DGS and can align the features of 3DGS well with the image and text spaces. More visualization results are available in the supplementary materials.

\subsection{Zero-Shot 3D Classification}
\label{sec:zs}

Leveraging the well-aligned multi-modal representations, CLIP-GS is naturally suited for zero-shot 3D classification.

\begin{table*}[t]
\scriptsize
  \setlength{\tabcolsep}{5pt}
  \centering
  
  \resizebox{0.9\textwidth}{!}{
    \begin{tabular}{lcccccccc}
    \toprule
    \multirow{2}[2]{*}{Method} & \multirow{2}[2]{*}{training data} & \multirow{2}[2]{*}{3D repr}  & \multicolumn{3}{c}{Objaverse-GS} & \multicolumn{3}{c}{ ModelNet-GS}\\
\cmidrule{4-9}          &&  & Top1  & Top3  & Top5  & Top1  & Top3  & Top5  \\
    \midrule
    PointCLIP \cite{zhang2022pointclip} & \multirow{2}[1]{*}{ShapeNet}  & PC \& Image & -  & -   & - & 23.8  &  - & -  \\
    PointCLIP v2 \cite{zhu2023pointclip}  &  & PC \& Image & -  &  -    & -  & 64.2  &     -  & - \\
    \midrule
    ULIP-PointBERT \cite{ulip} & \multirow{3}[4]{*}{Ensemble} & PC & -  & -   &- & 71.4  & 84.4  & 89.2  \\
    ULIP-2 \cite{ulip2} &  & PC & 31.1&	48.0&	55.4 & 75.6  & -  & 93.7  \\
    OpenShape-SparseConv \cite{openshape}  & \multirow{3}[2]{*}{(no LVIS)} & PC & 37.0  &   58.4    & 66.9  & 82.6  &   95.0    & 97.5 \\
     OpenShape-PointBERT \cite{openshape} & & PC & 39.1 & 60.8 & 68.9 & {85.3} & 96.2 & 97.4  \\
     Uni3D \cite{uni3d} & & PC & 47.7 & 69.8 & 77.3 & 86.2 & 97.4 &  \textbf{98.6} \\
     \textbf{CLIP-GS (ours)} & & 3DGS & \textbf{48.5} & \textbf{70.3} & \textbf{77.5} & \textbf{86.7} & \textbf{97.6} & \textbf{98.6} \\
    \midrule
    ULIP-PointBERT \cite{ulip}& \multirow{4}[2]{*}{Ensemble} & PC & -  &-    & -     & 75.1  & 88.1      & 93.2     \\
    OpenShape-SparseConv \cite{openshape} &       & PC & 43.4 & 64.8 & 72.4 & 83.4 & 95.6 & 97.8 \\
    OpenShape-PointBERT \cite{openshape} & & PC & {46.8} & {69.1} & {77.0} & 84.4 & {96.5} & {98.0} \\
     Uni3D \cite{uni3d} & & PC & 52.8 & 74.9 & 81.4 & 86.5 & 96.4 & 97.9  \\
     
     \textbf{CLIP-GS (ours)} & & 3DGS & \textbf{53.5} & \textbf{76.1} & \textbf{82.0} & \textbf{87.0} & \textbf{97.9} & \textbf{98.8}  \\

    \bottomrule
    \end{tabular}
    }
  \caption{Zero-shot classification on Objaverse-GS, and ModelNet-GS. ``no LVIS'' denotes model is trained without Objaverse-LVIS shapes.}
  \label{tab:zero-cls}%
  
\end{table*}%

\noindent \textbf{Dataset.}
We conduct experiments under two datasets:  Objaverse-GS and ModelNet-40 \cite{modelnet}. 
ModelNet-40 is a widely-used dataset that contains 40 common categories.
We used the ModelNet test dataset, which consists of 2,468 shapes, and generated the corresponding 3DGS using the method in Sec. \ref{sec:data} to construct the ModelNet-GS dataset. 

\noindent \textbf{Evaluation metrics.}
We follow the settings of \cite{uni3d, openshape}, using Top1, Top3, Top5 average accuracy ($\%$) for evaluations.

\noindent \textbf{Comparisons with state-of-the-art methods.}
We compare CLIP-GS with the previous state-of-the-art zero-shot 3D classification methods, including PointCLIP \cite{zhang2022pointclip, zhu2023pointclip}, ULIP \cite{ulip, ulip2}, OpenShape \cite{openshape} and Uni3D \cite{uni3d}.
PointCLIP projects point clouds into images and leverages CLIP for classification. ULIP, OpenShape, and Uni3D train 3D encoders to align the visual-text representation and use point clouds for classification. In contrast, Our CLIP-GS employs 3DGS as the input for classification. The results are shown in Tab. \ref{tab:zero-cls}. For a fair comparison, we present the results of Uni3D-Base, a 3D encoder model with $\sim$ 88M parameters.

CLIP-GS demonstrates a comprehensive improvement over existing zero-shot 3D classification models, achieving a performance boost of + 0.8, + 0.5 on Objaverse-GS and ModelNet-GS, respectively. Notably, we only use $\sim$ 240K 3DGS samples for training, yet the model exhibits excellent zero-shot capabilities. This data volume is far less than the million scales used for point cloud pre-training \cite{uni3d, openshape, ulip} and the billion scales used for image pre-training \cite{clip, evaclip}.

\subsection{Few-Shot 3D Classification}
\label{sec:fs}

\begin{table*}[t]
  \centering
  \hspace{-10.5mm}
  \begin{minipage}{.45\textwidth}
    \centering
	\begin{tabular}{l c }
	\toprule
		\multirow{1}*{Method}  &\multicolumn{1}{c}{~~~Acc. \& Dev.~~~}\\
		 \cmidrule(lr){1-1} \cmidrule(lr){2-2}
		DGCNN~\cite{wang2019dynamic} &86.3\ $\pm$\ 6.2 \\
		  DGCNN + OcCo~\cite{wang2021unsupervised} &86.4\ $\pm$\ 5.4 \\
		\cmidrule(lr){1-2}
	    PointTransformer~\cite{zhao2021point} &84.6\ $\pm$\ 5.5 \\
		 PointTransformer + OcCo~\cite{zhao2021point}~~~  &89.4\ $\pm$\ 5.1 \\
		 Point-BERT~\cite{yu2022point} &91.0\ $\pm$\ 5.4 \\
         Point-M2AE ~\cite{zhang2022point} &92.3\ $\pm$\ 4.5\\
        \midrule
        PointMamba~\cite{liang2024pointmamba} & 91.4\ $\pm$\ 4.4\\
        Mamba3D~\cite{han2024mamba3d}  &92.4\ $\pm$\ 4.1 \\
        PointRWKV~\cite{he2024pointrwkv} &94.8\ $\pm$\ 2.8  \\
        \midrule
	     \rowcolor{gray!10}\textbf{CLIP-GS (ours)} &\textbf{95.4\ $\pm$\ 0.2} \\
	\bottomrule
	\end{tabular}
    \caption{\textbf{Few-shot classification on ModelNet40}. We report the 10-shot \& 10-way average accuracy (\%) and standard deviation (\%) results. }
    \label{tab:fewshot1}
  \end{minipage}%
  \hspace{5.5mm}
  \begin{minipage}{.55\textwidth}
    \centering
        \renewcommand\arraystretch{1.05} 
	\begin{tabular}{lc c c c c}
	\toprule
		\multirow{2}*{Method} &\multicolumn{4}{c}{5-shot}\\
		 \cmidrule(lr){2-5} 
		 &5-way  &10-way  &20-way &50-way \\
		 \cmidrule(lr){1-1} \cmidrule(lr){2-5}
	    ULIP-2~\cite{ulip2} &90.5 &	85.0 &	80.0 &	71.5\\
	    OpenShape-SparseConv~\cite{openshape} &92.4 &	87.7 &	82.5 &	73.4\\
	    OpenShape-PointBERT~\cite{openshape} &92.0 &	88.0&	83.2 & 75.8\\
	    Uni3D~\cite{uni3d} &95.1  &92.2 &88.5 &82.1\\
		\textbf{CLIP-GS (ours)} & \textbf{95.6}&\textbf{92.6}&\textbf{89.2} &\textbf{82.5}\\
        \midrule
	    OpenShape-SparseConv*~\cite{openshape} &93.2&	88.9 &	83.9 &	75.4 \\
	    OpenShape-PointBERT*~\cite{openshape} &93.8 &	90.2 &	86.3 &	79.6\\
        Uni3D*~\cite{uni3d} &96.4 &94.1 &91.1  &85.6 \\
        \textbf{CLIP-GS* (ours)} & \textbf{96.6}  &\textbf{94.2}  &\textbf{91.4}  &\textbf{85.8}  \\
	\bottomrule
	\end{tabular}
    \caption{\textbf{Few-shot classification on Objaverse-GS}. We report the average accuracy (\%) for 5-shot classification across 5, 10, 20, and 50 ways. * denotes Objacerse-LVIS shapes are used during training.}
    \label{tab:fewshot2}
  \end{minipage}
\end{table*}

To evaluate the performance of CLIP-GS with limited data, we conduct experiments on few-shot 3D classification.

\noindent \textbf{Dataset \& task settings.}
We use ModelNet-GS and Objaverse-GS to conduct few-shot 3D classification benchmarks. 
Following \cite{he2024pointrwkv}, we use a $n$-way, $m$-shot setting, where $n$ denotes the number of classes randomly sampled from the dataset, and $m$ is the number of samples randomly drawn from each class. We experiment with $m =10$ and $n = 10$ in ModelNet-GS, and $m = 5$ and $n \in$ \{5, 10, 20, 50\} in Objaverse-GS. We do not construct 10-shot experiments on the Objaverse-GS since some classes contain fewer than 10 samples in Objaverse-GS.

\noindent \textbf{Evaluation metrics.}
In line with \cite{he2024pointrwkv}, we measure performance using Top1 average accuracy and standard deviation, with deviations calculated over 5 independent experiments.

\noindent \textbf{Comparisons with state-of-the-art methods.}
We compare CLIP-GS with the previous few-shot point-cloud classification methods \cite{wang2019dynamic, wang2021unsupervised, yu2022point, zhang2022point, liang2024pointmamba, han2024mamba3d, he2024pointrwkv} in ModelNet-GS (Tab. \ref{tab:fewshot1}). 
CLIP-GS surpasses previous state-of-the-art point cloud methods, and demonstrates significantly smaller deviations. Indicating CLIP-GS has a superior 3D encoding capability, enabling it to effectively cluster features of the same class and differentiate features of different classes.
We also compare CLIP-GS with the previous state-of-the-art multimodal point-cloud methods \cite{ulip2, uni3d, openshape} in Objaverse-GS (Tab. \ref{tab:fewshot2}).  
CLIP-GS consistently outperforms all the other methods under the few-shot settings of Objaverse-GS.

\subsection{Ablation Study}
\label{sec:abs}

We conduct ablation studies on various choices of designs within our CLIP-GS, and showcase their contributions to the final performance in Tab. \ref{tab:ab-comp}, \ref{tab:ab-order}, \ref{tab:ab-weight}, \ref{tab:ab-psnr}. 

\begin{table}[ht]
  \centering
    \begin{tabular}{l|cccc}
      \Xhline{0.7pt}

      & \textbf{3D repr} & \textbf{Top1} & \textbf{Top3} & \textbf{Top5} \\ 
      \hline
      Uni3D & $P \& C$ & 33.6 & 52.3 &  60.1  \\     
      \hline
      \multirow{2}[1]{*}{+ Fine-tune} & $P \& C$ & 46.9 & 68.5 & 75.9 \\ 
       & 3DGS & 44.8 & 66.3 & 74.1 \\ 
      \hline
      + GS Tokenizer & 3DGS & 47.9  & 69.9 &  76.8  \\     
      \hline
      \rowcolor{gray!10}\multicolumn{1}{c|}{+ Image Voting Loss} & 3DGS & \textbf{48.5} & \textbf{70.3} & \textbf{77.5}  \\     
      \Xhline{0.7pt}
      \end{tabular}
  \caption{Ablation of diverse designs of CLIP-GS. We use the Objaverse-GS for analysis. $P \& C$ denotes only $P$ and $C$ attributes of gaussian points from 3DGS is used.}
      \label{tab:ab-comp}
  \end{table}

\noindent \textbf{Component-wise ablations.}
To understand the effect of each component in the CLIP-GS, we start with the official pre-trained Uni3D and gradually add each design. (Tab. \ref{tab:ab-comp}). 
\begin{itemize}[itemsep=2pt,topsep=0pt,parsep=0pt]
\item \textbf{Baseline:} We use the point cloud-based method, Uni3D \cite{uni3d}, as the baseline model ($1^{st}$ row), and extract the $P$ and $C$ attributes of gaussian points from 3DGS to simulate the input format of the point cloud. Uni3D yields inferior performance. This result is due to the differing spatial placement between point cloud and 3DGS. \textit{i.e.}, point cloud is positioned on the surface of objects, whereas in 3DGS, gaussian points can exist within the object.
\item \textbf{Fine-tune:} Fine-tuning Uni3D on $P \& C$ resulting in an improvement of +13.3 Top1, +16.2 Top3, and +15.8 Top5 accuracy ($2^{nd}$ row). However, when using all 3DGS attributes, the TOP1 accuracy drops to 44.8 ($3^{rd}$ row), indicating that directly introducing these additional attributes can impair the network's learning.
\item \textbf{GS Tokenizer:} Using GS Tokenizer produce decent performance (the $4^{th}$ result), 33.6$\rightarrow$47.9 Top1, 52.3$\rightarrow$69.9 Top3, and 60.1$\rightarrow$76.8 Top5 accuracy.
\item \textbf{Image Voting Loss:} Finally, introducing image voting loss learns effective 3DGS and image alignment representation, further enhancing performance to establish state-of-the-art benchmarks (last row).
\end{itemize}

\begin{figure}[ht]
\begin{center}
  \vspace{-2mm}
   \includegraphics[width=0.99\linewidth]{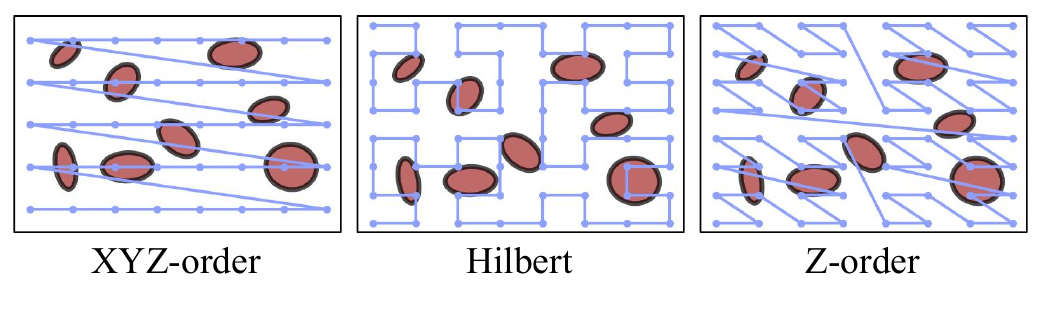}
\end{center}
  \vspace{-8mm}
   \caption{
    Visualization of different order strategies. We project the 3D space onto a 2D plane. 
   }
\label{fig:order}
\end{figure}

\noindent \textbf{Order indicator}
We investigate various strategies to reorder GS patches, including \textit{xyz}-order, Hilbert curve \cite{hilbert1935stetige}, and Z-order (details in Fig. \ref{fig:order}). For the \textit{xyz}-order, we used the x-axis order for visualization. The results are shown in Tab. \ref{tab:ab-order}.
Using one single ordering strategy leads to suboptimal results. Therefore, we employ a multiple-ordering strategy for collaborative ordering.
\begin{table}[ht]
  \centering
    \begin{tabular}{l|ccc}
      \Xhline{0.7pt}

       & \textbf{Top1} & \textbf{Top3} & \textbf{Top5} \\ 
      \hline
      \textit{xyz}-order & 48.3 & 70.4 &  77.6  \\     
      Hilbert & 48.4 & 70.5 &  77.7 \\     
      Z-order & 48.3  & 70.5 &  77.6  \\     
      \rowcolor{gray!10}\multicolumn{1}{c|}{all} & \textbf{48.5} & \textbf{70.3} & \textbf{77.5}  \\     
      \Xhline{0.7pt}
      \end{tabular}
  \caption{Ablation of the order indicator. Objaverse-GS is used. }
      \label{tab:ab-order}
  \vspace{-5mm}
  \end{table}

\noindent \textbf{Effect of pre-initialized weights.}
We conduct ablation studies on pre-initialized weights in Tab.~\ref{tab:ab-weight}, exploring the effectiveness of initializing transformer layers in CLIP-GS with either 2D pretraining models or point cloud pretraining models. We report the performance of training CLIP-GS from the 2D pretraining model EVA-CLIP \cite{evaclip} and the point cloud pretraining model Uni3D \cite{uni3d}. 
Using a point cloud pretraining weight for initialization presents significant performance advantages  (1$^{st}$ and 3$^{rd}$ row). We also experimented with freezing the parameters of the point cloud pretraining model and only training the GS Tokenizer, which resulted in diminished performance (2$^{nd}$ row).
\begin{table}[ht]
  \centering
    \begin{tabular}{l|ccc}
      \Xhline{0.7pt}

       & \textbf{Top1} & \textbf{Top3} & \textbf{Top5} \\ 
      \hline
      from 2D Image & 33.9  & 55.9 &  64.2  \\     
     from 3D PC (frozen) & 46.6  & 68.6 & 75.7  \\     
      \rowcolor{gray!10}\multicolumn{1}{l|}{from 3D PC} & \textbf{48.5} & \textbf{70.3} & \textbf{77.5}  \\     
      \Xhline{0.7pt}
      \end{tabular}
  \caption{Ablation of the pre-initialized weights in transformer layers. We use the Objaverse-GS for analysis. }
      \label{tab:ab-weight}
  \end{table}

\noindent \textbf{Restruction of 3DGS}
We analyzed the reconstruction quality (PSNR, SSIM), optimization cost (optimization time per 3D shape), and storage cost (average 3DGS storage size) of the 3DGS reconstruction process across different iterations and spherical harmonics (SH) degrees (Fig. \ref{tab:ab-psnr}). When SH degree=0, increasing iterations results in only modest improvements in reconstruction quality (+0.5 PSNR and +0.2 SSIM), while significantly increasing the training cost ($\times$ 3.7). Although a higher SH degree enhances the reconstruction quality, it also leads to a $\times$ 3.8 fold increase in storage demand. Therefore, we opt for 5000 iterations and an SH degree of 0, accepting a slight decrease in reconstruction quality in exchange for a $\times$ 3.9 increase in optimization speed and a $\times$3.6 reduction in storage cost.



\begin{table}[ht]
  \centering
  \footnotesize
    \renewcommand\arraystretch{1.2} 
    \begin{tabular}{cc|cccc}
      \Xhline{0.7pt}

       \textbf{Iter} & \textbf{SH} & \textbf{PSNR} & \textbf{SSIM} & \textbf{storage size} & \textbf{optimization time}\\ 
      \hline
      20,000 & 3 & 37.1 & 98.2 & 3.6M & 108.3s \\     
      20,000 & 0 &  35.1 & 97.9  & 1.0M & 104.5s\\     
      5,000 & 3 & 35.9 & 98.0 & 3.8 M & 29.9s \\     
      \rowcolor{gray!10}5,000 & 0 &  34.6 & 97.7 & 1.0 M & 28.1s\\     
      \Xhline{0.7pt}
      \end{tabular}
  \caption{Ablation of the reconstruction process of 3DGS, we record the model reconstruction quality (PSNR \& SSIM), average 3DGS storage size (MB) and optimization time (s/3dgs) at different iterations and SH degrees.}
      \label{tab:ab-psnr}
  \end{table}

\subsection{Scaling up model size}
\label{sec:scaling}
\begin{table}[ht]
  \centering
  \footnotesize
    \renewcommand\arraystretch{1.1} 
  \resizebox{0.5\textwidth}{!}{
    \begin{tabular}{c|ccc|cc}
      \Xhline{0.7pt}

       \textbf{Model name} & \textbf{Depth} & \textbf{Width} & \textbf{Heads} & \textbf{\#Params} & \textbf{Top1}\\ 
      \hline
      CLIP-GS-T & 12 & 192 & 3 & 7.3M & 45.9 \\     
      CLIP-GS-S & 12 & 384 & 6 & 23.7M & 47.0  \\     
      CLIP-GS-B & 12 & 768 & 12 & 89.8 M & 48.5 \\     
      CLIP-GS-L & 24 & 1024 & 16 & 308.2 M & 48.8 \\     
      \Xhline{0.7pt}
      \end{tabular}
      }
  \caption{Scaling up model size in CLIP-GS. Top1 accuracy in Objaverse-GS is used for analysis. }
      \label{tab:scale}
  \end{table}  

We explore the effect of scaling up the model size in Tab. \ref{tab:scale}. Following the scaling-up guidelines of \cite{uni3d}, we increase the model parameters from 7M (Tiny), 24M (Small), 89M (Base) to 308M (Large).
The results across different model scales consistently indicate that enlarging the model size of CLIP-GS enhances the 3D representation capabilities.
\section{Conclusion}
In this paper, we introduce CLIP-GS, a multimodal representation learning framework that aligns language, images, and 3DGS into a unified feature space. 
We also explore an efficient approach for generating 3DGS, rendered images, and text triplets.
CLIP-GS achieves state-of-the-art performance across various 3D perception tasks including multimodal retrieval, zero-shot 3D classification, and few-shot 3D classification. We hope CLIP-GS will serve as a solid baseline and help ease future research of 3D multimodal learning and related areas.

\noindent \textbf{Acknowledgment.} This work was funded by the Fundamental Research Funds for the Central Universities (2024XKRC082), the National Natural Science Foundation of China (No.92470203, U24B20179, 6120106009), and the Beijing Natural Science Foundation (No. L242022).

{
    \small
    \bibliographystyle{ieeenat_fullname}
    \bibliography{main}
}

\end{document}